\documentclass[runningheads]{llncs}

\usepackage{graphicx}
\usepackage{comment}
\usepackage{amsmath,amssymb} 
\usepackage{color}

\usepackage{float}
\usepackage[ruled, vlined, longend]{algorithm2e}
\usepackage{algpseudocode}
\usepackage{subfigure}
\usepackage{booktabs,multirow}
\usepackage{makecell}
\usepackage{dblfloatfix}
\usepackage[en-US]{datetime2}
\usepackage{hyperref}
\usepackage{fancyvrb}

\newcommand{\mywrt}[1]{\, \mathrm{d} #1}

\newcommand{\exptn}[1]{\mathbb{E} \left[ #1\right]}

\newcommand{\abs}[1]{\left\lvert#1\right\rvert}
\newcommand{\norm}[1]{\left\lVert#1\right\rVert}

\makeatletter
\newcommand{\printfnsymbol}[1]{%
	\textsuperscript{\@fnsymbol{#1}}%
}
\makeatother

\begin{document}
	\pagestyle{headings}
	\mainmatter
	\def\ECCVSubNumber{4234}  
	
	\title{Disentangling Multiple Features in Video Sequences using Gaussian Processes in Variational Autoencoders}
	
	
	
	\titlerunning{MGP-VAE}
	%
	\author{Sarthak Bhagat\inst{1} \and
		Shagun Uppal\inst{1}\printfnsymbol{1} \and
		Zhuyun Yin\inst{2} \and
		Nengli Lim\inst{3}
	}
	\authorrunning{S. Bhagat et al.}
	%
	\institute{IIIT Delhi \\
		\email{\{sarthak16189, shagun16088\}@iiitd.ac.in}
		\and
		Bioinformatics Institute, A*STAR, Singapore \\
		\email{yinzhuyun@gmail.com}
		\and
		Singapore University of Technology and Design \\
		\email{nengli\_lim@sutd.edu.sg}} 
	\maketitle

\begin{abstract}
We introduce MGP-VAE (Multi-disentangled-features Gaussian Processes Variational AutoEncoder), a variational autoencoder which uses Gaussian processes (GP) to model the latent space for the unsupervised learning of disentangled representations in video sequences. We improve upon previous work by establishing a framework by which multiple features, static or dynamic, can be disentangled. Specifically we use fractional Brownian motions (fBM) and Brownian bridges (BB) to enforce an inter-frame correlation structure in each independent channel, and show that varying this structure enables one to capture different factors of variation in the data. We demonstrate the quality of our representations with experiments on three publicly available datasets, and also quantify the improvement using a video prediction task. Moreover, we introduce a novel geodesic loss function which takes into account the curvature of
the data manifold to improve learning. Our experiments show that the combination of the improved representations with the novel loss function enable MGP-VAE to outperform the baselines in video prediction.
\end{abstract}

   \section{Introduction}
   Finding good representations for data is one of the main goals of unsupervised machine learning \cite{Bengio2012RepresentationLA}. Ideally, these representations reduce the dimensionality of the data, and are structured such that the different factors of variation in the data get distilled into different channels. This process of disentanglement in generative models is useful as in addition to making the data interpretable, the disentangled representations can also be used to improve downstream tasks such as prediction. \par 
   
   In prior work on the unsupervised learning of video sequences, a fair amount of effort has been devoted to separating motion, or dynamic information from static content \cite{Denton2017UnsupervisedLO,Grathwohl2016DisentanglingSA,Hsieh2018LearningTD,Li2018DisentangledSA,Villegas2017DecomposingMA}. To achieve this goal, typically the model is structured to consist of dual pathways, e.g. using two separate networks to separately capture motion and semantic content \cite{Denton2017UnsupervisedLO,Villegas2017DecomposingMA}. \par 

   Such frameworks may be restrictive as it is not immediately clear how to extend them to extract multiple static and dynamic features. Furthermore, in complex videos, there usually is not a clear dichotomy between motion and content, e.g. in videos containing dynamic information ranging over different time-scales. \par 

   In this paper, we address this challenge by proposing a new variational autoencoder, MGP-VAE (Multi-disentangled-features Gaussian Processes Variational AutoEncoder), for the unsupervised learning of video sequences. It utilizes a latent prior distribution that consists of multiple channels of fractional Brownian motions and Brownian bridges. By varying the correlation structure along the time dimension in each channel to pick up different static or dynamic features, while maintaining independence between channels, MGP-VAE is able to learn multiple disentangled factors. \par
   
   We then demonstrate quantitatively the quality of our disentanglement representations using a frame prediction task. To improve prediction quality, we also employ a novel geodesic loss function which incorporates the manifold structure of the data to enhance the learning process. \par 
   
   \begin{figure} 
   \centering  
    \includegraphics[height=4.5cm, width=\linewidth]{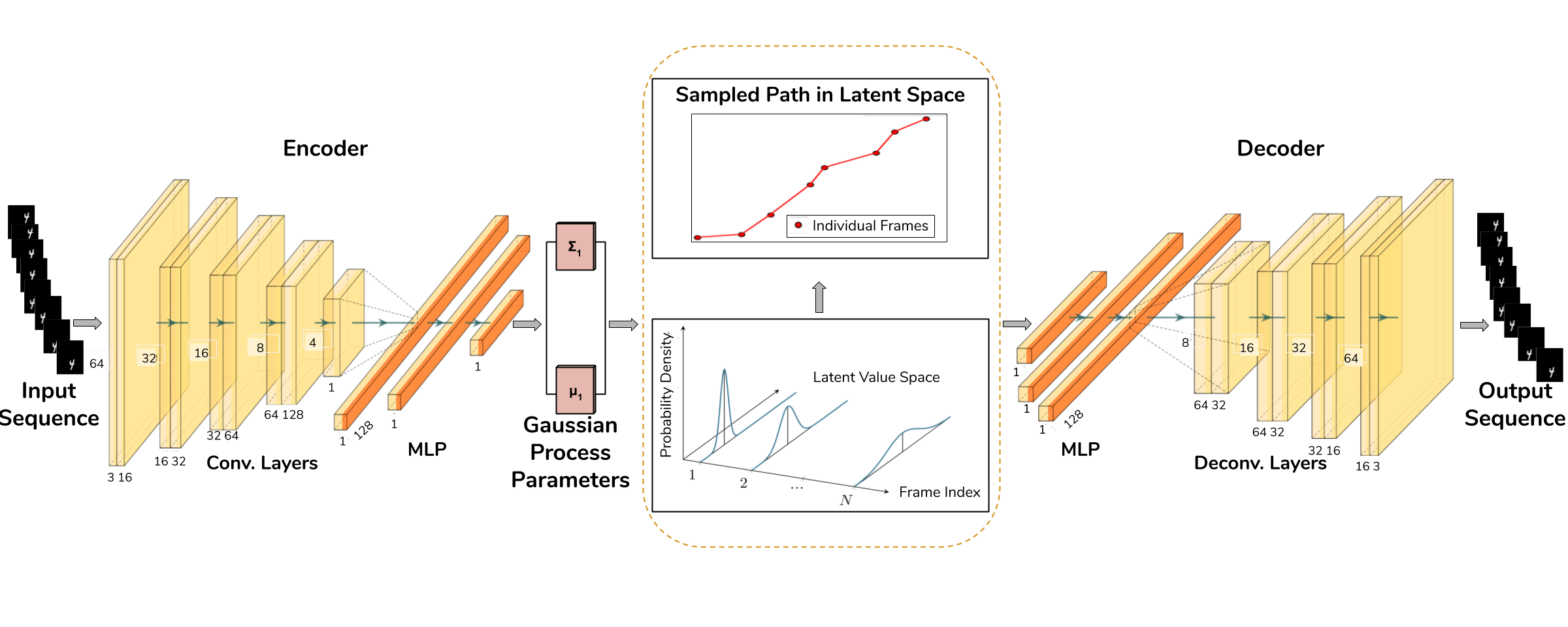}
   \caption{Network illustration of MGP-VAE: The network takes in a video sequence, an array of images, and encodes a Gaussian process latent space representation. The output of the encoder is the mean and covariance matrix of the Gaussian process, after which a sequence of points in $\mathbb{R}^d$ is sampled where each point represents one frame.}
   \end{figure}

   Our main contributions can be summarized as follows:
   \begin{itemize}
   \item We use Gaussian processes as the latent prior distribution in our model MGP-VAE to obtain disentangled representations for video sequences. Specifically, we structure the latent space by varying the correlation between video frame distributions so as to extract multiple factors of variation from the data. 
   \item We introduce a novel loss function which utilizes the structure of the data manifold to improve prediction. In particular, the actual geodesic distance between the predicted point and its target on the manifold is used instead of squared-Euclidean distance in the latent space.
   \item We test MGP-VAE against various other state-of-the-art models in video sequence disentanglement. We conduct our experiments on three datasets and use a video prediction task to demonstrate quantitatively that our model outperforms the competition.
   \end{itemize}
   
   \section{Related Work}
   
   \subsection{Disentangled Representation Learning for Video Sequences}
    There are several methods for improving the disentanglement of latent representations in generative models. InfoGAN \cite{infoGAN} augments generative adversarial networks \cite{GANs} by additionally maximizing the mutual information between a subset of the latent variables and the recognition network output. beta-VAE \cite{Higgins2017betaVAELB} adds a simple coefficient ($\beta$) to the KL divergence term in the evidence lower bound of a VAE. It has been demonstrated that increasing $\beta$ beyond unity improves disentanglement, but also comes with the price of increased reconstruction loss  \cite{Kim2018DisentanglingBF}. To counteract this trade-off, both FactorVAE \cite{Kim2018DisentanglingBF} and $\beta$-TCVAE \cite{tcVAE} further decompose the KL divergence term, and identify a total correlation term which when penalized directly encourages factorization in the latent distribution. \par 

   With regard to the unsupervised learning of sequences, there have been several attempts to separate dynamic information from static content \cite{Denton2017UnsupervisedLO,Grathwohl2016DisentanglingSA,Hsieh2018LearningTD,Li2018DisentangledSA,Villegas2017DecomposingMA}. In \cite{Li2018DisentangledSA}, one latent variable is set aside to represent content, separate from another set of variables used to encode dynamic information, and they employ this graphical model for the generation of new video and audio sequences.

   \cite{Villegas2017DecomposingMA} proposes MCnet, which uses a convolutional LSTM for encoding motion and a separate CNN to encode static content. The network is trained using standard $l_2$ loss plus a GAN term to generate sharper frames. DRNet \cite{Denton2017UnsupervisedLO} adopts a similar architecture, but uses a novel adversarial loss which penalizes semantic content in the dynamic pathway to learn pose features. \par 
   
   \cite{Hsieh2018LearningTD} proposes DDPAE, a model with a VAE structure that performs decomposition on video sequences with multiple objects in addition to disentanglement. In their experiments, they show quantitatively that DDPAE outperforms MCnet and DRNet in video prediction on the Moving MNIST dataset. \par 
   
   Finally, it has been shown that disentangled representation learning can be placed in the framework of nonlinear ICA \cite{Kingma2020}, particularly in the context of time-varying data \cite{Hyvarinen2016}.
   
   \subsection{VAEs and Gaussian Process Priors} 
   In \cite{Grathwohl2016DisentanglingSA}, a variational auto-encoder which structures its latent space distribution into two components is used for video sequence learning. The ``slow" channel extracts static features from the video, and the ``fast" channel captures dynamic motion. Our approach is inspired by this method, and we go further by giving a principled way to shape the latent space prior so as to disentangle multiple features. \par 
   
   Outside of video analysis, VAEs with a Gaussian process prior have also been explored. In \cite{Casale2018GaussianPP}, they propose GPPVAE and train it on image datasets of different objects in various views. The latent representation is a function of an object vector and a view vector, and has a Gaussian prior imposed on it. They also introduce an efficient method to speed up computation of the covariance matrices. \par 
   
   In \cite{Fortuin2019MultivariateTS}, a deep VAE architecture is used in conjunction with a Gaussian process to model correlations in multivariate time series such that inference can be performed on missing data-points. \par 
   
   Bayes-Factor VAE \cite{Kim2019BayesFactorVAEHB} uses a hierarchical Bayesian model to extend the VAE. As with our work, they recognize the limitations of restricting the latent prior distribution to standard normal, but they adopt heavy-tailed distributions as an alternative rather than Gaussian processes.

   \subsection{Data Manifold Learning}
   Recent work has shown that distances in latent space are not representative of the true distance between data-points \cite{Arvanitidis2017LatentSO,Khnel2018LatentSN,Shao2017TheRG}. Rather, deep generative models learn a mapping from the latent space to the data manifold, a smoothly varying lower-dimensional subset of the original data space. \par 

   In \cite{shapeAnalysis}, closed curves are abstractly represented as points on a shape manifold which incorporates the constraints of scale, rotational and translational invariance. The geodesic distance between points on this manifold is then used to give an improved measure of dissimilarity.   In \cite{Shukla2019GeometryOD}, several metrics are proposed to quantify the curvature of data manifolds arising from VAEs and GANs. 
   
   \section{Method}
   In this section, we review the preliminaries on VAEs and Gaussian processes, and describe our model MGP-VAE in detail.

   \subsection{VAEs}
   
   Variational autoencoders \cite{Kingma2013AutoEncodingVB} are powerful generative models which reformulate autoencoders in the framework of variational inference. Given latent variables $z \in \mathbb{R}^M$, the decoder, typically a neural network, models the generative distribution $p_{\theta}(x \,|\, z)$, where $x \in \mathbb{R}^N$ denotes the data. Due to the intractability of computing the posterior distribution $p(z \,|\, x)$, an approximation $q_{\phi}(z \,|\, x)$, again parameterized by another neural network called the encoder, is used. Maximizing the log-likelihood of the data can be achieved by maximizing the evidence lower bound
   \begin{align}
   \mathbb{E}_{q_{\phi} \left(z | x \right)} \left[ \log \frac{p_{\theta}(x, z)}{q_{\phi}(z \, \big| \, x)} \right],
   \end{align}
   which is equal to
   \begin{align} \label{lossFn}
   \mathbb{E}_{q_{\phi} \left(z | x \right)} \left[ \log p_{\theta}(x | z) \right] - D_{KL} \left[ q_{\phi} \left(z | x \right) \, \Big| \, p(z) \right],
   \end{align}
   with $p(z)$ denoting the prior distribution of the latent variables. \par 

   The negative of the first term in \eqref{lossFn} is the reconstruction loss, and can be approximated by
   \begin{align}
   \frac{1}{L} \sum_{l=1}^L -\log p^{\vphantom{0}}_{\theta} \left( x \, \big| \, z^{(l)} \right),
   \end{align}   
   where $z^{(l)}$ is drawn ($L$ times) from the latent distribution, although typically only one sample is required in each pass as long as the batch size is sufficiently large \cite{Kingma2013AutoEncodingVB}. If $p^{\vphantom{0}}_{\theta} \left( x \, \big| \, z \right)$ is modeled to be Gaussian, then this is simply mean-squared error.
         
   \subsection{Gaussian Processes}
   Given an index set $T$, $\left\{ X_t; t\in T \right\}$ is a Gaussian process \cite{hidaGP,williams2006gaussian} if for any finite set of indices $\left\{t_1,...,t_n \right\}$ of $T$, $\left(X_{t_1},...,X_{t_n} \right)$ is a multivariate normal random variable. In this paper, we are concerned primarily in the case where $T$ indexes time, i.e. $T = \mathbb{R}^+$ or $\mathbb{Z}^+$, in which case $\left\{ X_t; t\in T \right\}$ can be uniquely characterized by its mean and covariance functions
   \begin{align}
   &\mu(t) := \exptn{X_t}, \\
   &R(s, t) := \exptn{X_t X_s}, \quad \forall \, s, t \in T.
   \end{align}
   
   The following Gaussian processes are frequently encountered in stochastic models, e.g. in financial modeling \cite{bfg2016,glasserman}, and the prior distributions employed in MGP-VAE will be the appropriately discretized versions of these processes.

    \begin{figure}
   \centering  
   \includegraphics[height = 4.2cm, width = 10cm]{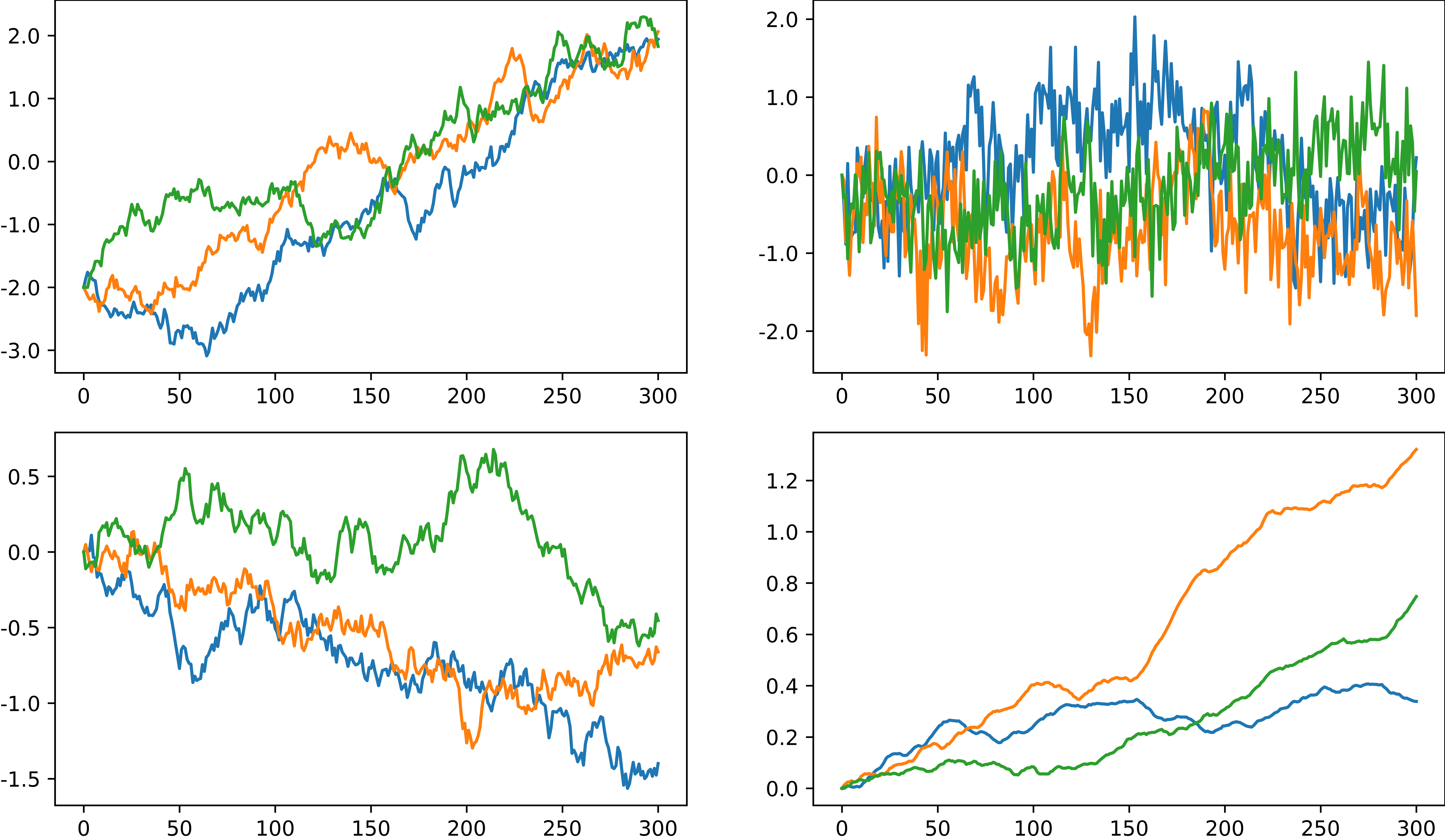}
   \caption{Sample paths for various Gaussian processes. Top-left: Brownian bridge from -2 to 2; top-right: fBM with H = 0.1; bottom-left: standard Brownian motion; bottom-right: fBM with H = 0.9}
   \end{figure}

\noindent \textbf{Fractional Brownian Motion (fBM). }
   Fractional Brownian motion \cite{fBMMandelbrot} $\left\{ B^H_t; t\in T \right\}$ is a Gaussian process parameterized by a Hurst parameter $H \in (0, 1)$, with mean and covariance functions given by
   \begin{align} \label{fbmParam}
   &\mu(t) = 0, \\
   &R(s, t) = \frac{1}{2} \left(s^{2H} + t^{2H} - \abs{t - s}^{2H} \right), \quad \forall \, s, t \in T.
   \end{align}
   When $H = \frac{1}{2}$, $W_t := B^{\frac{1}{2}}_t$ is standard Brownian motion \cite{hidaGP} with independent increments, i.e. the discrete sequence $(W^{\frac{}{}}_0, W_1, W_2, \ldots)$ is a simple symmetric random walk where $W_{n+1} \sim \mathcal{N} (W_n, 1)$. \par 
   Most notably, when $H \neq \frac{1}{2}$, the process is not Markovian. When $H > \frac{1}{2}$, the disjoint increments of the process are positively correlated, whereas when $H < \frac{1}{2}$, they are negatively correlated. We will demonstrate in our experiments how tuning $H$ effects the clustering of the latent code. \par
   
   \noindent \textbf{Brownian Bridge (BB). }
   The Brownian bridge \cite{glasserman,Karatzas1987BrownianMA} from $a \in \mathbb{R}$ to $b \in \mathbb{R}$ on the domain $[0, T]$ is the Gaussian process defined as
   \begin{align} \label{bb}
   X_t = a \left( 1 - \frac{t}{T} \right) + b \left( \frac{t}{T} \right) + W_t + \frac{t}{T} W_T.
   \end{align}
   Its mean function is identically zero and its covariance function is given by 
   \begin{align} \label{BBCov}
   R(s, t) = \min(s, t) - \frac{st}{T}, \quad \forall \, s, t \in T.
   \end{align}
   It can be also represented as the solution to the stochastic differential equation \cite{Karatzas1987BrownianMA}
   \begin{align}
   \mathrm{d}X_t = \frac{b - X_t}{T-t} \mywrt{t} + \mywrt{W_t}, \quad X_0 = a,
   \end{align}
   with solution
   \begin{align}
   X_t = a \left( 1 - \frac{t}{T} \right) + b \left( \frac{t}{T} \right) + (T-t) \int_0^t \frac{1}{T-s} \mywrt{W_s}.
   \end{align}
   From \eqref{bb}, its defining characteristic is that it is pinned at the start and the end such that $X_0 = a$ and $X_T = b$ almost surely.
   
   \subsection{MGP-VAE} \label{section:sampling}
   For VAEs in the unsupervised learning of static images, the latent distribution $p(z)$ is typically a simple Gaussian distribution, i.e. $z \sim \mathcal{N}(0, \sigma^2 \mathcal{I}_d)$. For a video sequence input $(x_1, \ldots x_n)$ with $n$ frames, we model the corresponding latent code as   
   \begin{align}
   &z = (z_1, z_2, \ldots, z_n) \sim \mathcal{N} (\mu_0, \Sigma_0), \quad z_i \in \mathbb{R}^d, \\ 
   &\mu_0 = \left[ \mu_0^{(1)}, \ldots, \mu_0^{(d)} \right] \in \mathbb{R}^{n \times d}, \\
   &\Sigma_0 = \left[ \Sigma_0^{(1)}, \ldots, \Sigma_0^{(d)} \right] \in \mathbb{R}^{n \times n \times d}.
   \end{align}
   Here $d$ denotes the number of channels, where one channel corresponds to one sampled Gaussian path, and for each channel, $\left\{ \mu_0^{(i)}, \Sigma_0^{(i)} \right\}$ are the mean and covariance of 
   \begin{align}
   V + \sigma B^H_t, \quad t = \{1, \ldots, n\},
   \end{align}
   in the case of fBM or
   \begin{align}
   A \left( 1 - \frac{t}{n} \right) + B \left( \frac{t}{n} \right) + \sigma \left( W_t + \frac{t}{n} W_n \right)
   \end{align}
   in the case of Brownian bridge. $V$, $A$ are initial distributions, and $B$ is the terminal distribution for Brownian bridge. They are set to be standard normal, and we experiment with different values for $\sigma$. The covariances can be computed using \eqref{fbmParam} and \eqref{BBCov} and are not necessarily diagonal, which enables us to model more complex inter-frame correlations. \par 

   Rewriting $z$ as $\left( z^{(1)}, \ldots, z^{(d)} \right)$, for each channel $i = 1, \ldots, d$, we sample $z^{(i)} \in \mathbb{R}^n \sim \mathcal{N} \left( \mu_0^{(i)}, \Sigma_0^{(i)} \right)$ by sampling from a standard normal $\xi$ and computing
   \begin{align}
   z^{(i)} = \mu_0^{(i)} + L^{(i)} \xi,
   \end{align}
   where $L^{(i)}$ is the lower-triangular Cholesky factor of $\Sigma_0^{(i)}$. \par 

   The output of the encoder is a mean vector $\mu_1$ and a symmetric positive-definite matrix $\Sigma_1$, i.e.
   \begin{align}
   q(z \,|\, x) \sim \mathcal{N} (\mu_1, \Sigma_1),
   \end{align}
   and to compute the KL divergence term in \eqref{lossFn}, we use the formula
   \small 
   \begin{align}
   D_{KL} \left[ q \, | \, p \right] 
   &= \frac{1}{2} \bigg[ \mathrm{tr} \left( \Sigma_0^{-1} \Sigma_1 \right) + \left\langle \mu_1 - \mu_0, \Sigma_0^{-1} (\mu_1 - \mu_0) \right\rangle - k + \log \left(\frac{\det \Sigma_1}{\det \Sigma_0}\right) \bigg].
   \end{align}
   \normalsize
   Following \cite{Higgins2017betaVAELB}, we add a $\beta$ factor to the KL divergence term to improve disentanglement. We will describe the details of the network architecture of MGP-VAE in Section 4.

   \subsection{Video Prediction Network and Geodesic Loss Function} 

   For video prediction, we predict the last $k$ frames of a sequence given the first $n-k$ frames as input. To do so, we employ a simple three-layer MLP (16 units per layer) with ReLU activation which operates in latent space rather than on the actual frame data so as to best utilize the disentangled representations. The first $n-k$ frames are first encoded by a pre-trained MGP-VAE into a sequence of points in latent space. These points are then used as input to the three-layer MLP to predict the next point, which is then passed through MGP-VAE's decoder to generate the frame. This process is then repeated $k-1$ more times. \par 
   
	Given an output $z_0$ and a target $z_T$, we use the geodesic distance between $g(z_0)$ and $g(z_T)$ as the loss function instead of the usual squared-distance $\norm{z_0 - z_T}^2$. Here, $g: \mathbb{R}^{n \times d} \rightarrow M \subset \mathbb{R}^N$ is the differentiable map from the latent space to the data manifold $M$ which represents the action of the decoder. We use the following algorithm from \cite{Shao2017TheRG} to compute the geodesic distance.
   
   \begin{algorithm} [H]
      \SetAlgoLined
      \textbf{Input: } Two points, $z_{0}, z_{T} \in Z$; \\
      $\alpha$, the learning rate \\
      \textbf{Output: } Discrete geodesic path, $z_{0}, z_{1}, ..., z_{T} \in Z$ \\
      Initialize $z_{i}$ as the linear interpolation between $z_{0}$ and $z_{T}$  \\
      \While{$\Delta E_{z_t} > \epsilon$}
      {
         \For{$i \in \{1, 2, ... ,T-1\}$}{
            Compute gradient using \eqref{eq:energy_gradient} \\
            $z_{i} \leftarrow z_{i} - \alpha\nabla_{z_t} E_{z_t} $
         }
      }
      \caption{Geodesic Interpolation}
      \label{algo:geodesic_interpolation}
   \end{algorithm}

   This algorithm finds the minimum of the energy of the path (and thus the geodesic)
   \begin{align}
   E_{z_t} = \frac{1}{2} \sum_{i=0}^T \frac{1}{\delta t} \norm{g(z_{i+1}) - g(z_i)}^2
   \end{align}
   by computing its gradient
   \begin{align}
   \nabla_{z_t} E_{z_t} = - \left( \nabla g (z_i) \right)^T \left[g(z_{i+1}) - 2g(z_i) + g(z_{i-1}) \right].
   \label{eq:energy_gradient}
   \end{align}

\begin{figure} [H]
      \centering  
      \includegraphics[height = 3.875cm, width = 8cm]{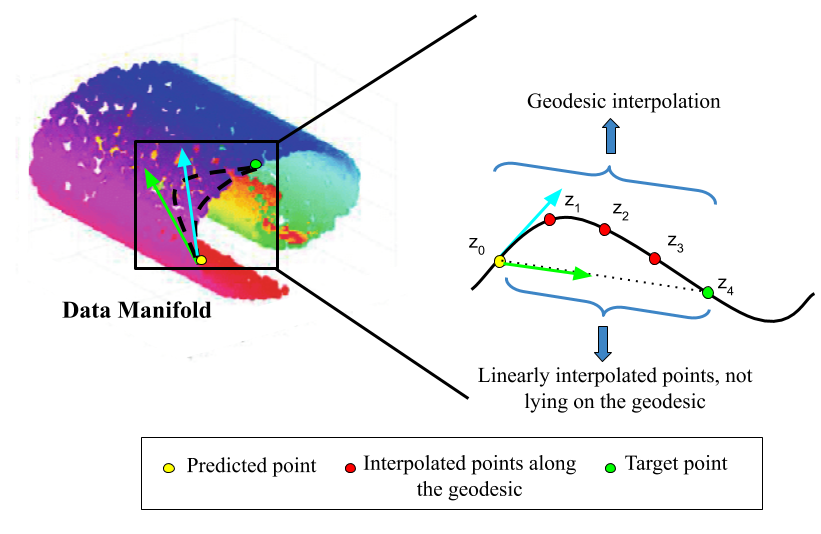}      
      \caption{Using the geodesic loss function as compared to squared-distance loss for prediction. By setting the target as $z_1$ instead of $z_4$, the model learns more efficiently to predict the next point.}
      \label{fig:geodesic_loss}
   \end{figure}

	Algorithm \ref{algo:geodesic_interpolation} initializes $\{z_i\}$ to be uniformly-spaced points along the line between $z_0$ and $z_T$ and gradually modifies them until the change in energy falls below a predetermined threshold. At this point, we use $z_1$ as the target instead of $z_T$ as $z_1 - z_0$ is more representative of the vector in which to update the prediction $z_0$ such that the geodesic distance is minimized; see Figure \ref{fig:geodesic_loss} for an illustration. \par
    
   \section{Experiments}
   In this section, we present experiments which demonstrate MGP-VAE's ability to disentangle multiple factors of variation in video sequences. \par
   
   \subsection{Datasets}
    
    \textbf{Moving MNIST}\footnote{\href{http://www.cs.toronto.edu/~nitish/unsupervised\_video}{http://www.cs.toronto.edu/~nitish/unsupervised\_video}} \cite{mmnist} comprises of moving gray-scale hand-written digits. We generate 60,000 sequences for training, each with a single digit moving in a random direction across frames and bouncing off edges. \par
   
    \noindent \textbf{Coloured dSprites} is a modification of the dSprites\footnote{\href{https://github.com/deepmind/dsprites-dataset}{https://github.com/deepmind/dsprites-dataset}} \cite{Higgins2017betaVAELB} dataset. It consists of 2D shapes (square, ellipse, heart) with 6 values for scale and 40 values for orientation. We modify the dataset by adding 3 variations for colour (red, green, blue) and constrain the motion of each video sequence to be simple horizontal or vertical motion. \par 

   For each sequence, the scale is set to gradually increase or decrease a notch in each frame. Similarly, after an initial random selection for orientation, the subsequent frames rotate the shape clockwise or anti-clockwise one notch per frame. The final dataset consists of a total of approximately 100,000 datapoints. \par

   \noindent \textbf{Sprites} \cite{Reed2015DeepVA} comprises of around 17,000 animations of synthetically rendered animated caricatures. There are 7 attributes: body type, sex, hair type, arm type, armor type, greaves type, and weapon type, with a total of 672 unique characters. In each animation, the physical traits of the sprite remain constant while the pose (hand movement, leg movement, orientation) is varied. \par
   
   \subsection{Network Architecture and Implementation Details}
   For the encoder, we use 8 convolutional layers with batch normalization between each layer. The number of filters begins with 16 in the first layer and increases to a maximum of 128 in the last layer. An MLP layer follows the last layer, and this is followed by another batch normalization layer. Two separate MLP layers are then applied, one which outputs a lower-triangular matrix which represents the Cholesky factor of the covariance matrix of $q(z \,|\,x)$ and the other outputs the mean vector. \par 
   
   For the decoder, we have 7 deconvolutional layers, with batch normalization between each layer. The first layer begins with 64 filters and this decreases to 16 filters by the last layer. We use ELU for the activation functions between all layers to ensure differentiability, with the exception of the last layer, where we use a hyperbolic tangent function. \par 
   
   Table \ref{parameters} lists the settings for the manually tuned hyperparameters in the experiments. All channels utilizing Brownian bridge (BB) are conditioned to start at $-2$ and end at $2$. \par
	
  \begin{table} [H]
   \scriptsize
      \begin{center}
     	\caption{Hyperparameter settings for all datasets}
      	\label{parameters}
         \begin{tabular}{ccccc} \hline 
            & \textbf{Moving MNIST} & \textbf{Coloured dSprites} & \textbf{Sprites} \\ \hline
            \thead{\textbf{Gaussian} \\ \textbf{processes}} & \makecell{Channel 1: fBM (H = 0.1) \\ Channel 2: fBM (H = 0.9)} & 5 Channels of BBs & 5 Channels of BBs \\
            \hline 
            \textbf{$\sigma$} & 0.25 & 0.25 & 0.25 \\
            \textbf{$\beta$} & 2 & 2 & 2 \\
            \textbf{Learning Rate} & 0.001 & 0.008 & 0.010 \\
            \textbf{No. of epochs} & 200 & 120 & 150 \\ \hline 
         \end{tabular}
      \end{center}

   \end{table}
   
   \subsection{Qualitative Analysis}
      
\begin{figure} [H]
    \centering
   \subfigure{
   \includegraphics[scale=0.6]{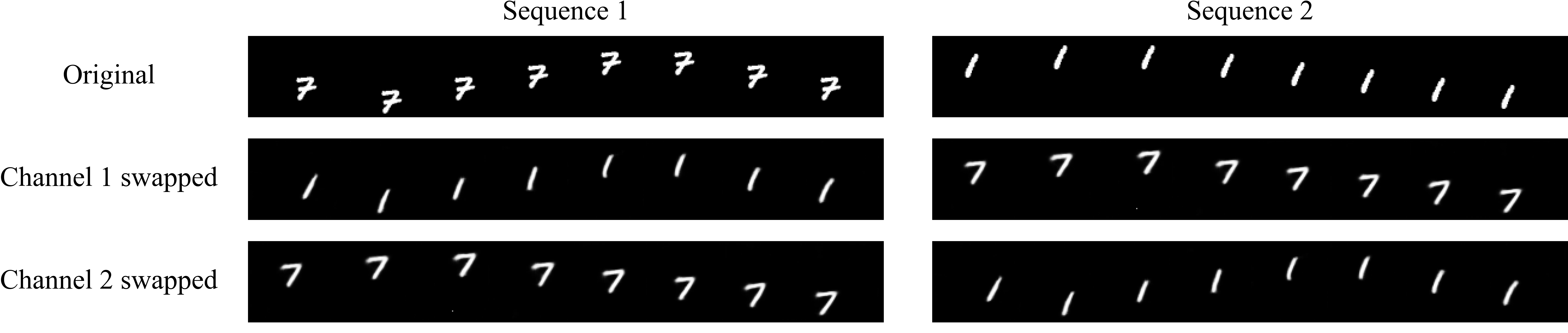}}
   \subfigure{
   \includegraphics[scale=0.6]{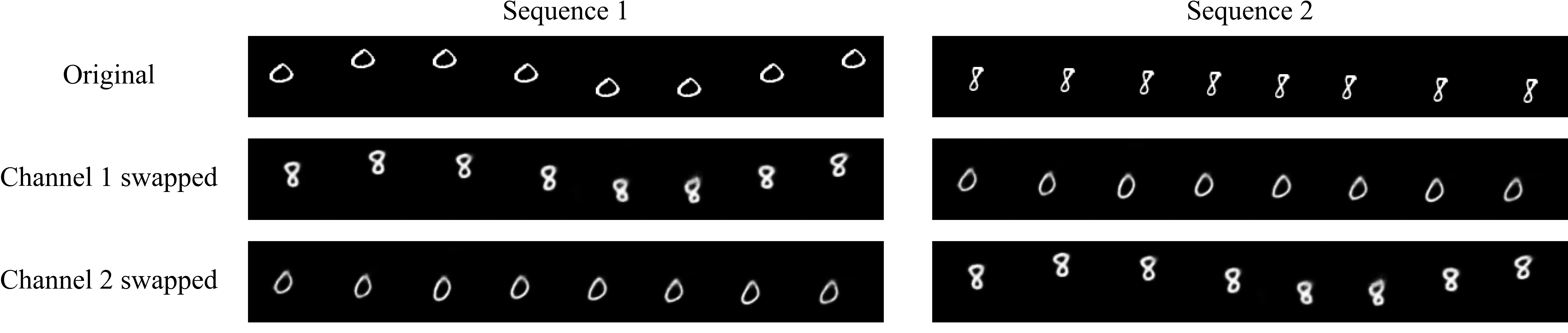}}
   \caption{Results from swapping latent channels in Moving MNIST; channel 1 (fBM(\(H=0.1\))) captures digit identity; channel 2 (fBM(\(H=0.9\))) captures motion.}
   \label{bmnist}
   \end{figure}

   Figure \ref{bmnist} shows the results from swapping latent channels in the Moving MNIST dataset, where we see that channel 1 (fBM(\(H=0.1\))) captures the digit identity, whereas channel 2 (fBM(\(H=0.9\))) captures the motion. \par 

\begin{figure} [H]
   \centering  
   \subfigure[fBM, H = 0.1]{
   \includegraphics[scale=0.35]{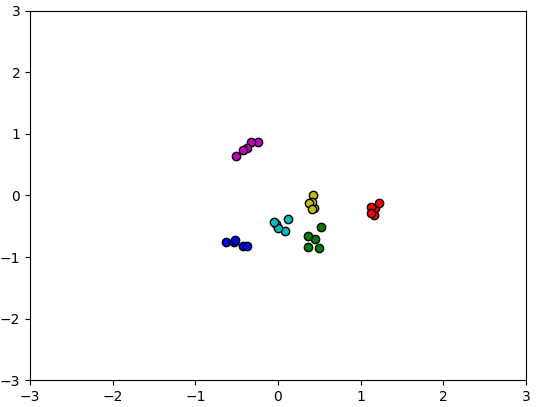}} 
   \subfigure[fBM, H = 0.9]{
   \includegraphics[scale=0.35]{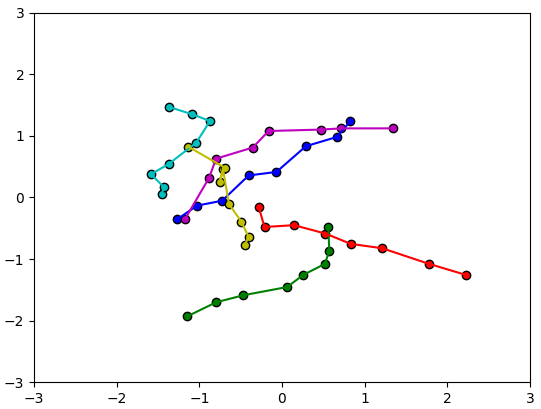}}
   \caption{Latent space visualization of fBM channels for 6 videos. Each point represents one frame of a video. The more tightly clustered points in (a) capture digit identity whereas the scattered points in (b) capture motion.}
   \label{fig:latent_vis}
   \end{figure}

   Figure \ref{fig:latent_vis} gives a visualization of the latent space (here we use two channels with $H = 0.1$ and two channels with $H = 0.9$). In our experiments, we observe that fBM channels with $H = 0.9$ are able to better capture motion in comparison to setting $H = 0.5$ (simple-symmetric random walk, cf. \cite{Grathwohl2016DisentanglingSA}). We hypothesize that shifting the value of $H$ away from that of the static channel sets the distributions apart and allows for better disentanglement. \par

   \begin{figure} [H]
   \centering  
   \subfigure{
   \includegraphics[scale=0.5]{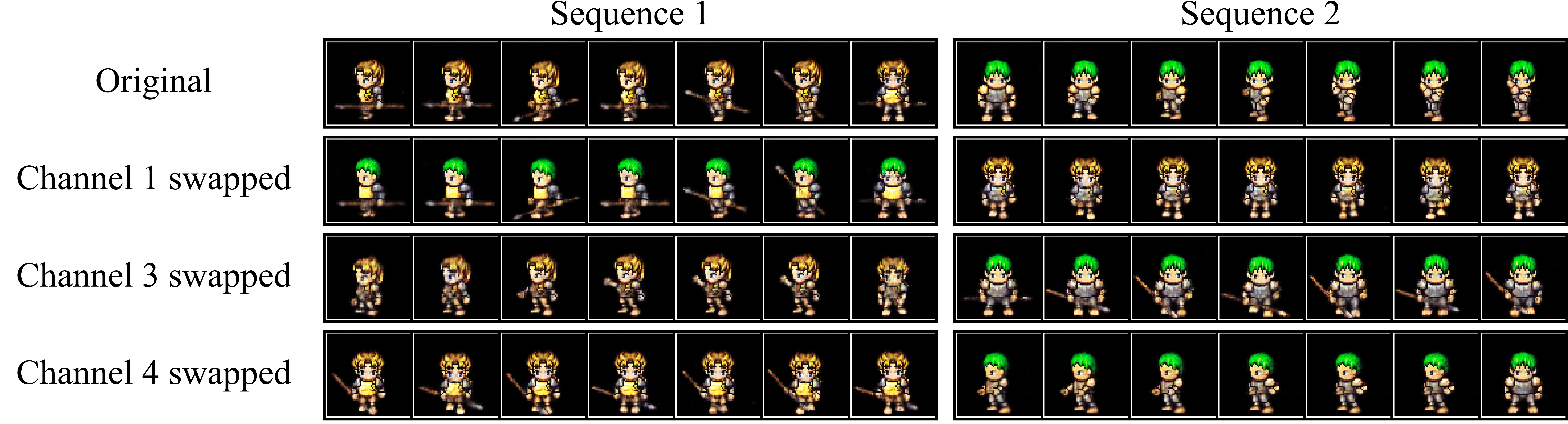}}
   \subfigure{
   \includegraphics[scale=0.5]{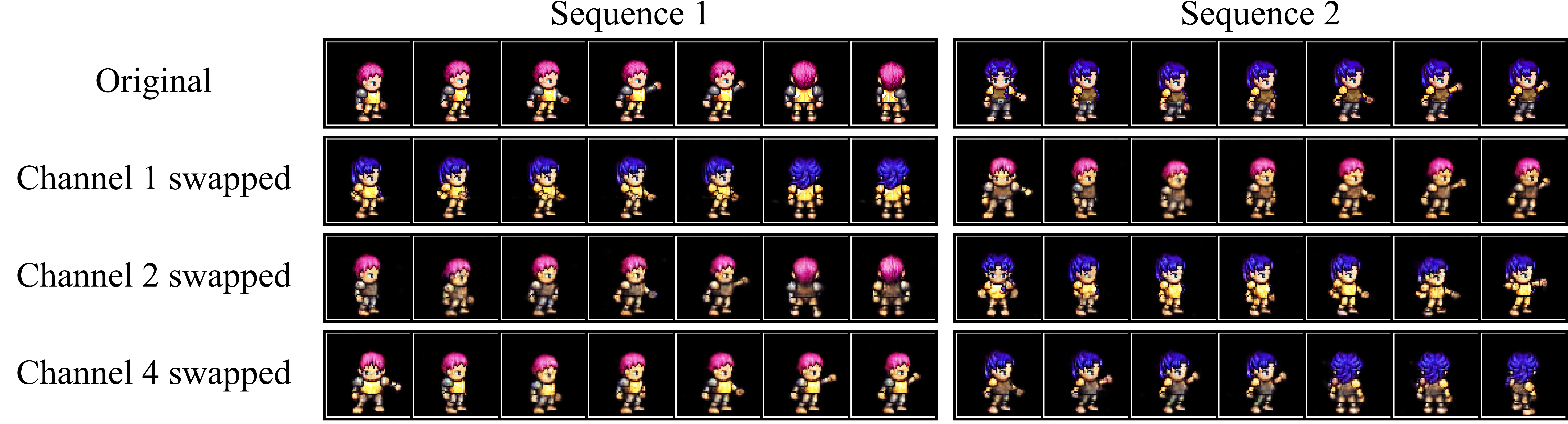}}
   \caption{Results from swapping latent channels in Sprites; channel 1 captures hair type, channel 2 captures armor type, channel 3 captures weapon type, and channel 4 captures body orientation.}
   \label{sprites}
   \end{figure}

\begin{figure} [H]
   \centering  
   \subfigure{
   \includegraphics[scale=0.45]{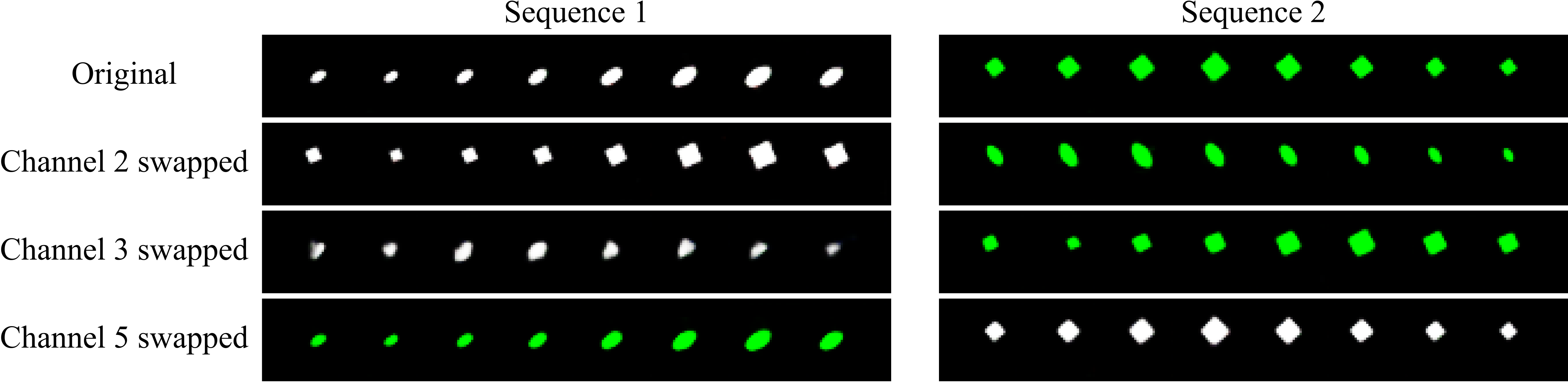}
   }
   \subfigure{
      \includegraphics[scale=0.45]{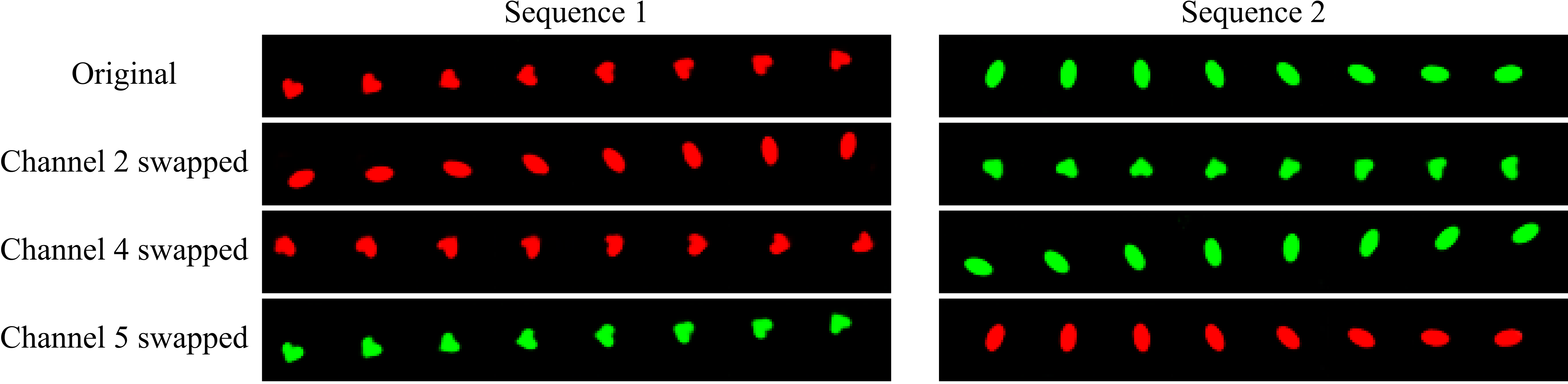}
      }
   \caption{Results from swapping latent channels in Coloured dSprites; channel 2 captures shape, channel 3 captures scale, channel 4 captures orientation and position, and channel 5 captures color.}
   \label{cdsprites}
   \end{figure}

   Figures \ref{sprites} and \ref{cdsprites} show the results from swapping latent channels in the Sprites dataset and Coloured dSprites dataset respectively. \par 
         
   \noindent \textbf{Discussion. }
   The disentanglement results were the best for Moving MNIST, where we achieved full disentanglement in more than 95\% of the cases. We were also able to consistently disentangle three or more features in Coloured dSprites and Sprites, but disentanglement of four or more features occurred less frequently due to their complexity. \par 
   
   We found that including more channels than the number of factors of variation in the dataset improved disentanglement, even as the extra channels did not necessarily encode anything new. For the Coloured dSprites and Sprites dataset, we originally experimented with different combinations of fBMs (with varying $H$) and Brownian bridges, but found that simply using 4-5 channels of Brownian bridges gave comparable results. We observed that with complex videos not easily separated into static or dynamic content, incorporating multiple Brownian bridge channels each with different start and end points led to good disentanglement. We hypothesize that anchoring the start and end points of the sequence at various places in latent space ``spreads out" and improves the representation. \par 

   Finally, we also tested other Gaussian processes such as the Ornstein-Ulenbeck process \cite{OUProcess} but as the results were not satisfactory, we shall defer a more detailed investigation to future work. \par 

   \subsection{Evaluating Disentanglement Quality}
   We first evaluate the disentangled representations by computing the mean average precision of a k-nearest neighbor classification over labeled attributes in the Coloured dSprites and Sprites datasets. 


	\begin{table}[H]
	\scriptsize
      \centering
        \caption{mAP values (\%) for Coloured dSprites and Sprites}        
        \begin{tabular}{c|cccccc|ccccccc} \hline       
          \label{mAP}
          \multirow{2}{*}{\textbf{Model}} & \multicolumn{6}{c}{\textbf{Coloured dSprites}} & \multicolumn{5}{c}{\textbf{Sprites}} \\ 
          & Shape & Color & Scale & x-Pos & y-Pos & Avg. & Gender & Skin & Vest & Hair & Arm & Leg & Avg. \\ \hline
          MCnet & 95.6 & 94.0 & 69.2 & 69.7 & 70.2 & 79.7 & 78.8 & 70.8 & 76.6 & 80.2 & 78.2 & 70.7 & 75.9 \\ 
          DRNet & 95.7 & 94.8 & 69.6 & 72.4 & 70.6 & 80.6 & 80.5 & 70.8 & 77.0 & 78.6 & 79.7 & 71.4 & 76.3 \\ 
          DDPAE & 95.6 & 94.2 & 70.3 & 71.6 & 72.4 & 80.8 & 79.8 & 72.0 & 77.4 & 79.3 & 78.3 & 74.6 & 76.9 \\ 
          MGP-VAE & 96.2 & 94.0 & 77.9 & 76.4 & 72.8 & \textbf{83.4} & 80.3 & 71.8 & 76.8 & 82.3 & 79.9 & 79.8 & \textbf{78.5} \\ \hline
          \end{tabular}
	\end{table}

	Table \ref{mAP} shows that our model is able to capture multiple features more effectively than the baselines MCnet\footnote{\href{https://github.com/rubenvillegas/iclr2017mcnet}{https://github.com/rubenvillegas/iclr2017mcnet}} \cite{Villegas2017DecomposingMA}, DRNet\footnote{\href{https://github.com/ap229997/DRNET}{https://github.com/ap229997/DRNET}} \cite{Denton2017UnsupervisedLO} and DDPAE\footnote{\href{https://github.com/jthsieh/DDPAE-video-prediction}{https://github.com/jthsieh/DDPAE-video-prediction}} \cite{Hsieh2018LearningTD}. \par 

	Next, we use a non-synthetic benchmark in the form of a video prediction task to illustrate the improvement in the quality of MGP-VAE's disentangled representations. We train a prediction network with the geodesic loss function as outlined in Section 3.4, where we set the number of interpolated points to be four. In addition, to speed up the algorithm for faster training, we ran the loop in Algorithm \ref{algo:geodesic_interpolation} for a fixed number of iterations (10-15) instead of until convergence. \par
   
   We compute the pixel-wise mean-squared-error and binary cross-entropy between the predicted $k$ frames and the actual last $k$ frames, given the first $n-k$ frames as input ($n$ is set to 8 for Moving MNIST and Coloured dSprites, and set to 7 for Sprites). Tables \ref{mse_bce_mnist} and \ref{mse_bce_dsprites} below summarize the results. \par

   \begin{table}[H]
      \centering
      \caption{Prediction results on Moving MNIST}        
		\begin{tabular}{ccccc} \hline 
		    \label{mse_bce_mnist}
			& \multicolumn{2}{c}{$k = 1$} 
			& \multicolumn{2}{c}{$k = 2$}  \\ \cmidrule(lr){2-3} \cmidrule(lr){4-5}				
			\textbf{Model} & \textbf{MSE} & \textbf{BCE} & \textbf{MSE} & \textbf{BCE} \\ \hline					
			MCnet \cite{Villegas2017DecomposingMA} & 50.1 & 248.2 & 91.1 & 595.5 \\		
			DRNet \cite{Denton2017UnsupervisedLO} & 45.2 & 236.7 & 86.3 & 586.7 \\		
			DDPAE \cite{Hsieh2018LearningTD} & 35.2 & 201.6 & 75.6 & 556.2 \\		
			Grathwohl, Wilson \cite{Grathwohl2016DisentanglingSA} & 59.3 & 291.2 & 112.3 & 657.2 \\		
			MGP-VAE & 25.4 & 198.4 & 72.2 & 554.2 \\		
			MGP-VAE (with geodesic loss) & \textbf{18.5} & \textbf{185.1} & \textbf{69.2} & \textbf{531.4} \\ \hline
      \end{tabular}
	\end{table}

  \begin{table}[!htb]
      \centering
        \caption{Last-frame ($k = 1$) prediction results for Coloured dSprites and Sprites}        
        \begin{tabular}{ccccc} \hline       
          \label{mse_bce_dsprites}
          \textbf{Dataset} & \multicolumn{2}{c}{\textbf{Coloured dSprites}} 
          & \multicolumn{2}{c}{\textbf{Sprites}}  \\ \cmidrule(lr){2-3} \cmidrule(lr){4-5}          
          \textbf{Model} & \textbf{MSE} & \textbf{BCE} & \textbf{MSE} & \textbf{BCE} \\ \hline               
          MCnet \cite{Villegas2017DecomposingMA} & 20.2 & 229.5  & 100.3 & 2822.6 \\          
          DRNet \cite{Denton2017UnsupervisedLO} & 15.2 & 185.2 & 94.4 & 2632.1 \\       
          DDPAE \cite{Hsieh2018LearningTD} & 12.6 & 163.1 & 75.4 & 2204.1 \\      
          MGP-VAE & 6.1 & 85.2 & 68.8 & 1522.5 \\      
          MGP-VAE (with geodesic loss) & \textbf{4.5} & \textbf{70.3} & \textbf{61.6} & \textbf{1444.4} \\ \hline 
          \end{tabular}
\end{table}

   The results show that MGP-VAE \footnote{\href{https://github.com/SUTDBrainLab/MGP-VAE}{https://github.com/SUTDBrainLab/MGP-VAE}}, even without using the geodesic loss function, outperforms the other models. Using the geodesic loss functions further lowers MSE and BCE. DDPAE, a state-of-the-art model in video disentanglement, achieves comparable results, although we note that we had to train the model considerably longer on the Coloured dSprites and Sprites datasets as compared to Moving MNIST to get the same performance. \par 

   Using the geodesic loss function during the training of the prediction network also leads to qualitatively better results. Figure \ref{qual} below shows that in a sequence with large MSE and BCE losses, the predicted point can generate an image frame which differs considerably from the actual image frame when the normal loss function is used. This is rectified with the geodesic loss function.

   \begin{figure} [H]
      \centering      
      \includegraphics[height=1.25cm, width=5.0cm]{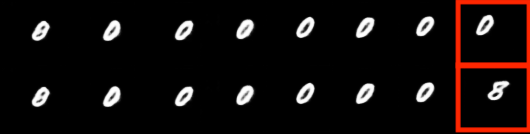}
      \includegraphics[height=1.25cm, width=5.0cm]{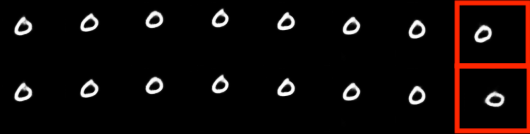}      
      \caption{Qualitative improvements from using the geodesic loss function: Left: without geodesic loss function; Right: with geodesic loss function; Top row: original video; Bottom row: video with the predicted last frame.}
      \label{qual}
  \end{figure}
   
   \section{Conclusion}
   We introduce MGP-VAE, a variational autoencoder for obtaining disentangled representations from video sequences in an unsupervised manner. MGP-VAE uses Gaussian processes, such as fractional Brownian motion and Brownian bridge, as a prior distribution for the latent space. We demonstrate that different parameterizations of these Gaussian processes allow one to extract different static and time-varying features from the data. \par 

   After training the encoder which outputs a disentangled representation of the input, we demonstrate the efficiency of the latent code by using it as input to a MLP for video prediction. We run experiments on three different datasets and demonstrate that MGP-VAE outperforms the baseline models in video frame prediction. To further improve the results, we introduce a novel geodesic loss function which takes into account the curvature of the data manifold. This contribution is independent of MGP-VAE, and we believe it can be used to improve video prediction in other models as well. \par 

   For future work, we will continue to experiment with various combinations of Gaussian processes. In addition, enhancing our approach with more recent methods such as FactorVAE, $\beta$-TCVAE, or independent subspace analysis  \cite{Stuehmer2020} may lead to further improvements.

\end{document}